\documentclass{article}

\usepackage{amssymb}
\usepackage{graphicx}
\usepackage{cite}
\usepackage{xcolor}
\usepackage{url}
\usepackage{graphicx}
\usepackage{amsmath}
\usepackage{multirow}
\usepackage{amsfonts}
\usepackage{amsthm}

\usepackage[ruled]{algorithm2e} 

\usepackage{algorithmic}
\usepackage{authblk}

\usepackage{textcomp}

\def\BibTeX{{\rm B\kern-.05em{\sc i\kern-.025em b}\kern-.08em
    T\kern-.1667em\lower.7ex\hbox{E}\kern-.125emX}}
\begin{document}

\title{k-Nearest Neighbors by Means of Sequence to Sequence Deep Neural Networks and Memory Networks}

\author[1]{Yiming Xu}
\author[1]{Diego Klabjan}

\affil[1]{Northwestern University}
\maketitle

\begin{abstract}
k-Nearest Neighbors is one of the most fundamental but effective classification models. In this paper, we propose two families of models built on a sequence to sequence model and a memory network model to mimic the k-Nearest Neighbors model, which generate a sequence of labels, a sequence of out-of-sample feature vectors and a final label for classification, and thus they could also function as oversamplers. We also propose `out-of-core' versions of our models which assume that only a small portion of data can be loaded into memory. Computational experiments show that our models on structured datasets outperform k-Nearest Neighbors, a feed-forward neural network, XGBoost, lightGBM, random forest and a memory network, due to the fact that our models must produce additional output and not just the label. On image and text datasets, the performance of our model is close to many state-of-the-art deep models. As an oversampler on imbalanced datasets, the sequence to sequence kNN model often outperforms Synthetic Minority Over-sampling Technique and Adaptive Synthetic Sampling.
\end{abstract}


\section{Introduction}


Recently, neural networks have been attracting a lot of attention among researchers in both academia and industry, due to their astounding performance in fields such as natural language processing \cite{nlp1}\cite{nlp2} and image recognition \cite{ir1}\cite{ir2}. Interpretability of these models, however, has always been an issue since it is difficult to understand the performance of neural networks. The well-known manifold hypothesis states that real-world high dimensional data (such as images) form lower-dimensional manifolds embedded in the high-dimensional space \cite{ir5}, but these manifolds are tangled together and are difficult to separate. The classification process is then equivalent to stretching, squishing and separating the tangled manifolds apart. However, these operations pose a challenge: it is quite implausible that only affine transformations followed by pointwise nonlinear activations are sufficient to project or embed data into representative manifolds that are easily separable by class. 

Therefore, instead of asking neural networks to separate the manifolds by a hyperplane or a surface, it is more reasonable to require points of the same manifold to be closer than points of other manifolds \cite{colah}. Namely, the distance between manifolds of different classes should be large and the distance between manifolds of the same class should be small. This distance property is behind the concept of k-Nearest Neighbor (kNN) \cite{knn}. Consequently, letting neural networks mimic kNN would combine the notion of manifolds with the desired distance property. 


We explore kNN through two deep neural network models: sequence to sequence deep neural networks \cite{s2s1} and memory networks \cite{MN14}\cite{MN15}. A family of our models are based on a sequence to sequence network. The new sequence to sequence model has the input sequence of length 1 corresponding to a sample, and then it decodes it to predict two sequences of output, which are the classes of closest samples and neighboring samples not necessarily in the training data, where we call the latter as out-of-sample feature vectors. We also propose a family of models built on a memory network, which has a memory that can be read and written to and is composed of a subset of training samples, with the goal of using it for predicting both classes of close samples and out-of-sample feature vectors. With the help of attention over memory vectors, our new memory network model generates the predicted label sequence and out-of-sample feature vectors. Both families of models use loss functions that mimic kNN. Computational experiments show that the new sequence to sequence kNN model consistently outperforms benchmarks (kNN\cite{knn}, random forest\cite{RF}, XGBoost\cite{XGB}, lightGBM\cite{lgbm}, a feed-forward neural network and a vanilla memory network) on structured datasets. The performance on some commonly used image and text datasets is comparable to many state-of-the-art deep models. We postulate that this is due to the fact that we are forcing the model to `work harder' than necessary (producing out-of-sample feature vectors).

Different from general classification models, our models predict not only labels, but also out-of-sample feature vectors. Usually a classification model only predicts labels, but as in the case of kNN, it is desirable to learn or predict the feature vectors of neighbors as well. Intuitively, if a deep neural network predicts both labels and feature vectors, it is forced to learn and capture representative information of input, and thus it should perform better in classification. Moreover, our models also function as synthetic oversamplers: we add the out-of-sample feature vectors and their labels (synthetic samples) to the training set. Experiments show that our sequence to sequence kNN model outperforms Synthetic Minority Over-sampling Technique (SMOTE) \cite{smote} and Adaptive Synthetic sampling (ADASYN) \cite{adasyn} most of the times on imbalanced datasets.

Usually we allow models to perform kNN searching on the entire dataset, which we call the full versions of models, but kNN is computationally expensive on large datasets. We design an algorithm to resolve this and we test our models under such an `out-of-core' setting: only a batch of data can be loaded into memory, i.e. kNN searching in the entire dataset is not allowed. For each such random batch, we compute the $K$ closest samples with respect to the given training sample. We repeat this $R$ times and find the closest $K$ samples among these $KR$ samples. These closest $K$ samples provide the approximate label sequence and feature vector sequence to the training sample based on the kNN algorithm. Computational experiments show that sequence to sequence kNN models and memory network kNN models significantly outperform the kNN benchmark under the out-of-core setting. 

Our main contributions are as follows. First, we develop two types of deep neural network models which mimic the kNN structure. Second, our models are able to predict both labels of closest samples and out-of-sample feature vectors at the same time: they are both classification models and oversamplers. Third, we establish the out-of-core version of models in the situation where not all data can be read into computer memory or kNN cannot be run on the entire dataset. The full version of the sequence to sequence kNN models and the out-of-core version of both sequence to sequence kNN models and memory network kNN models outperform the benchmarks, which we postulate is because learning neighboring samples enables the model to capture representative features. 

We introduce background and related works in Section II, show our approaches in Section III, and describe datasets and experiments in Section IV. Conclusions are in Section V.

\section{Background and Literature Review}

There are several works trying to mimic kNN or applying kNN within different models. \cite{bf} introduced the boundary forest algorithm which can be used for nearest neighbor retrieval. Based on the boundary forest model, in \cite{dbt}, a boundary deep learning tree model with differentiable loss function was presented to learn an efficient representation for kNN. The main differences between this work and our work are in the base models used (boundary tree vs standard kNN), in the main objectives (representation learning vs classification and oversampling) and in the loss functions (KL divergence vs KL divergence components reflecting the kNN strategy and $L^2$ norm). \cite{knntext} introduced a text classification model which utilized nearest neighbors of input text as the external memory to predict the class of input text. Our memory network kNN models differ from this model in 1) the external memory: our memory network kNN models simply feed a random batch of samples into the external memory without the requirement of nearest neighbors and thus they save computational time, and 2) number of layers: our memory network kNN models have $K$ layers while the model proposed by \cite{knntext} has one layer. A higher-level difference is that \cite{knntext} considered a pure classification setting, while our models generate not only labels but out-of-sample feature vectors as well. Most importantly, the loss functions are different: \cite{knntext} used KL divergence as the loss function while we use a specially designed KL divergence and $L^2$ norm to force our models to mimic kNN. 

The sequence to sequence model, one of our base models, has recently become the leading framework in natural language processing \cite{s2s1}\cite{gru}. In \cite{gru} an RNN encoder-decoder architecture was used to deal with statistical machine translation problems. \cite{s2s1} proposed a general end-to-end sequence to sequence framework, which is used as the basic structure in our sequence to sequence kNN model. The major difference between our work and these studies is that the loss function in our work forces the model to learn from neighboring samples, and our models are more than just classifiers - they also create out-of-sample feature vectors that improve accuracy or can be used as oversamplers. 

There are also a plethora of studies utilizing external memory or attention technique in neural networks. \cite{MN14} proposed the memory network model to predict the correct answer of a query by means of ranking the importance of sentences in the external memory. \cite{MN15} introduced a continuous version of a memory network with a recurrent attention mechanism over an external memory, which outperformed the previous discrete memory network architecture in question answering. Since it has shown strong ability of capturing long term dependencies of sequential data, our memory network kNN model is built based on the end-to-end memory network model. Moreover, \cite{att1} introduced an attention-based model to search for the informative part of input sentence to generate prediction. \cite{att2} proposed a word-by-word attention-based model to encourage reasoning about entailment. These works utilized an attention vector over inputs, but in our work, the attention is over the external memory rather than the input sequence of length 1.
 
In summary, the main differences between our work and previous studies are as follows. First, our models predict both labels of nearest samples and out-of-sample feature vectors rather than simply labels. Thus, they are more than classifiers: the predicted label sequences and feature vector sequences can be treated as synthetic oversamples to handle imbalanced class problems. Second, our work emphasizes on the out-of-core setting. 
All of the prior works related to kNN and deep learning assume that kNN can be run on the entire dataset and thus cannot be used on large datasets. 
Third, our loss functions are designed to mimic kNN, so that our models are forced to learn neighboring samples to capture the representative information. 


\subsection{Sequence to Sequence model}
A family of our models are built on sequence to sequence models. A sequence to sequence (Seq2seq) model is an encoder-decoder model. The encoder encodes the input sequence to an internal representation called the `context vector' which is used by the decoder to generate the output sequence. Usually, each cell in the Seq2seq model is a Long Short-Term Memory (LSTM) cell \cite{lstm2} or a Gated Recurrent Unit (GRU) cell \cite{gru}. 

Given input sequence $x_1,...,x_T$, in order to predict output $Y^P_1, ..., Y^P_{K}$ (where the superscript $P$ denotes `predicted'), the Seq2seq model estimates conditional probability $P(Y^P_1,...,$ $Y^P_t|x_1,...,x_T)$ for $1\leq t\leq K$. At each time step $t$, the encoder updates the hidden state $h^e_t$, which can also include the cell state, by 
$$
h^e_t=f^e_h(x_t,h^e_{t-1})
$$where $1\leq t\leq T$.
The decoder updates the hidden state $h^d_t$ by 
$$
h^d_t=f^d_h(Y^P_{t-1},h^d_{t-1},h^e_T)\\
$$ where $1\leq t\leq K$.  
The decoder generates output $y_t$ by 
\begin{equation}
y_t=g(Y^P_{t-1},h^d_t,h^e_T), 
\end{equation}
and
$$Y^P_t=q(y_t)
$$with $q$ usually being softmax function.

The model calculates the conditional probability of output $Y^P_1,...,Y^P_{K}$ by
$$\Pr(Y^P_1, ..., Y^P_{K}|x_1, ..., x_T)=\prod_{t=1}^{K}\Pr(Y^P_t|Y^P_1, ..., Y^P_{t-1}).$$


\subsection{End to End Memory Networks}

The other family of our models are built on an end-to-end memory network (MemN2N). This model takes $x_1,...,x_n$ as the external memory, a `query' $x$, a ground truth $Y^{GT}$ and predicts an answer $Y^P$. It first embeds memory vectors $x_1,...,x_n$ and query $x$ into continuous space. They are then processed through multiple layers to generate the output label $Y^P$. 

MemN2N has $K$ layers. In the $t^{th}$ layer, where $1\leq t\leq K$, the external memory is converted into embedded memory vectors $m^t_1,...,m^t_n$ by an embedding matrix $A^t$. The query $x$ is also embedded as $u^t$ by an embedding matrix $B^t$. The attention scores between embedded query $u^t$ and memory vectors $(m_i^t)_{i=1,2,...,n}$ are calculated by
$$p^t = softmax((u^t)^T m^t_1, (u^t)^T m^t_2, ..., (u^t)^T m^t_n).$$
Each $x_i$ is also embedded to an output representation $c^t_i$ by another embedding matrix $C^t$. The output vector from the external memory is defined as
$$o^t= \mathlarger{\sum}_{i=1}^n p^t_i c^t_i.
$$By a linear mapping $H$, the input to the next layer is calculated by
$$u^{t+1} = Hu^t + o^t.
$$\cite{MN15} suggested that the input and output embeddings are the same across different layers, i.e. $A^1 = A^2 = ... = A^K$ and $C^1 = C^2 = ... = C^K$. 
In the last layer, by another embedding matrix $W$, MemN2N generates a label for the query $x$ by
$$Y^P = softmax(W( Hu^K+o^K)).
$$

\section{kNN models}

Our sequence to sequence kNN models are built on a Seq2seq model, and our memory network kNN models are built on a MemN2N model. Let $K$ denote the number of neighbors of interest.

\subsection{Vector to Label Sequence (V2LS) Model}

Given an input feature vector $x$, a ground truth label $Y^{GT}$ (a single class corresponding to $x$) and a sequence of labels $Y^T_1,Y^T_2,...,Y^T_K$ corresponding to the labels of the $1^{st},2^{nd},...,K^{th}$ nearest sample to $x$ in the entire training set, V2LS predicts a label $Y^P$ and $Y^P_1,Y^P_2,...,Y^P_K$, the predicted labels of the $1^{st},2^{nd},...,K^{th}$ nearest samples. Since $Y^T_1,Y^T_2,...,Y^T_K$ are obtained by using kNN upfront, the real input is only $x$ and $Y^{GT}$. When kNN does not misclassify, $Y^{GT}$ corresponds to majority voting of $Y^T_1,Y^T_2,...,Y^T_K$.

The key concept of our model is to have $x$ as the input sequence (of length 1) and the output sequence $Y^P_1,Y^P_2,...,Y^P_K$ to correspond to $Y^T_1,Y^T_2,...,Y^T_K$. The loss function also captures $Y^{GT}$ and $Y^T_1,Y^T_2,...,Y^T_K$.

In the V2LS model, by a softmax operation with temperature after a linear mapping $(W_y,b_y)$, the label of the $t^{th}$ nearest sample to $x$ is predicted by
$$Y_t^P=softmax((W_yy_t+b_y)/\tau)$$
where $y^t$ is as in (1) for $t=1,2,...,K$ and $\tau$ is the temperature of softmax.   

By taking the average of predicted label distributions, the label of $x$ is predicted by
$$Y^P= \mathlarger{\sum}_{t=1}^{K}Y_t^P/K.$$
Note that if $Y^P_t$ corresponds to a Dirac distribution for each $t$, then $Y^P$ matches majority voting. 
Temperature $\tau$ controls the ``peakedness" of $Y^P_t$. Values of $\tau$ below 1 push $Y^P_t$ towards a Dirac distribution, which is desired in order to mimic kNN \cite{karp}\cite{temperature}. 
We design the loss function as
$$L_1=\displaystyle \mathop{\mathbb{E}_{}}\{\mathlarger{\sum}_{t=1}^{K}D_{KL}(Y_t^T||Y_t^P)/K+\alpha D_{KL}(Y^{GT}||Y^P)\}$$
where the first term captures the label at the neighbor level, the second term for the actual ground truth and $\alpha$ is a hyperparameter to balance the two terms. The expectation is taken over all training samples, and $D_{KL}$ denotes the Kullback-Leibler divergence. Due to the fact that the first term is the sum of KL divergence between predicted labels of nearest neighbors and target labels of nearest neighbors, it forces the model to learn information about the neighborhood. The second term considers the actual ground truth label: a classification model should minimize the KL divergence between the predicted label (average of $K$ distributions) and the ground truth label. By combining the two terms, the model is forced to not only learn the classes of the final label but also the labels of nearest neighbors. We let the $t^{th}$ decoder cell predict the $t^{th}$ nearest sample because the preceding decoder cells preserve closeness to the original input. In the subsequent cells, the information gets passed through more decoder cells and thus it is expected to deviate more from the input, which is the reason why we let the $t^{th}$ cell predict the $t^{th}$ nearest neighbor. Experiments in Section IV also show the importance of preserving the order of nearest neighbors. 

In inference, given an input $x$, V2LS predicts $Y^P$ and $Y^P_1,Y^P_2,...,Y^P_K$, but only $Y^P$ is the actual output; it is used to measure the classification performance. Note that it is possible that $argmax{Y^P}$ is different from the majority voted class among $argmax{Y^P_1},argmax{Y^P_2},...,$ $argmax{Y^P_K}$ when kNN misclassifies.

\subsection{Vector to Vector Sequence (V2VS) Model}

We use the same structure as the V2LS model except that in this model, the inputs are a feature vector $x$ and a sequence of feature vectors $X_1^T,X_2^T,...,X_K^T$ corresponding to the $1^{st},2^{nd},...,K^{th}$ nearest sample to $x$ among the entire training set (calculated upfront using kNN). V2VS predicts $X_1^P,X_2^P,...,X_K^P$ which denote the predicted out-of-sample feature vectors of the $1^{st},2^{nd},...,K^{th}$ nearest sample. Since $X_1^T,X_2^T,...,X_K^T$ are obtained using kNN, this is an unsupervised model.

The output of the $t^{th}$ decoder cell $y_t$ is processed by a linear layer ($W_{x1},b_{x1}$), a $ReLU$ operation and another linear layer ($W_{x2},b_{x2}$) to predict the out-of-sample feature vector
$$X_t^P=W_{x2}max\{W_{x1}y_t+b_{x1},0\}+b_{x2}, t=1,2,...,K.$$
Numerical experiments show that $ReLU$ works best compared with $tanh$ and other activation functions. The loss function is defined to be the sum of $L^2$ norms as
$$L_2=\displaystyle \mathop{\mathbb{E}_{}}\{\mathlarger{\sum}_{t=1}^{K}||X_t^P-X_t^T||^2\}.$$
Since the predicted out-of-sample feature vectors should be close to the input vector, learning nearest vectors forces the model to learn a sequence of approximations to something very close to the identity function. However, this is not trivial. First it does not learn an exact identity function, since the output is a sequence of nearest neighbors to input, i.e. it does not simply copy the input $K$ times. Second, by limiting the number of hidden units of the neural network, the model is forced to capture the most representative and condensed information of input. A large amount of studies have shown this to be beneficial to classification problems \cite{ae1}\cite{ae3}\cite{ae2} \cite{ae4}. 

In inference, we predict the label of $x$ by finding the labels of out-of-sample feature vectors $X_t^P$ and then perform majority voting among these $K$ labels.

\subsection{Vector to Vector Sequence and Label Sequence (V2VSLS) Model}

In previous models, V2LS learns to predict labels of nearest neighbors and V2VS learns to predict feature vectors of nearest neighbors. Combining V2LS and V2VS together, this model predicts both $X_t^P$ and $Y_t^P$. Given an input feature vector $x$, a ground truth label $Y^{GT}$, a sequence of nearest labels $Y^T_1,Y^T_2,...,Y^T_K$ and a sequence of nearest feature vectors $X^T_1,X^T_2,...,X^T_K$, V2VSLS predicts a label $Y^P$, a label sequence $Y^P_1,Y^P_2,...,Y^P_K$ and an out-of-sample feature vector sequence $X^P_1,X^P_2,...,X^P_K$. Since the two target sequences are obtained by kNN, the model still only needs $x$ and $Y^{GT}$ as input.

The loss function is a weighted sum of the two loss functions in V2LS and V2VS
$$L=L_1+\lambda L_2$$
where $\lambda$ is a hyperparameter to account for the scale of the $L^2$ norm and the KL divergence. 

The $L^2$ norm part enables the model to learn neighboring vectors. As discussed in the V2VS model, this is beneficial to classification since it drives the model to capture representative information of input and nearest neighbors. The $KL$ part of the loss function focuses on predicting labels of nearest neighbors. As discussed in the V2LS model, the two terms in the $KL$ loss force the model to learn both neighboring labels and the ground truth label. Combining the two parts, the V2VSLS model is able to predict nearest labels and out-of-sample feature vectors, as well as one final label for classification.

In inference, given an input $x$, V2VSLS generates $Y^P$, $X^P_1,X^P_2,...,X^P_K$ and $Y^P_1,Y^P_2,...,Y^P_K$. Still only $Y^P$ is used in measuring classification performance of the model.

\subsection{Memory Network - kNN (MNkNN) Model}

The MNkNN model is built on the MemN2N model, which has $K$ layers stacked together. After these layers, the MemN2N model generates a prediction. In order to mimic kNN, our MNkNN model has $K$ layers as well but it generates one label after each layer, i.e. after the $t^{th}$ layer, it predicts the label of the $t^{th}$ nearest sample. Similar to the Seq2seq kNN models, the $t^{th}$ layer predicts the $t^{th}$ nearest sample because the preceding layers preserve closeness to the input. Therefore, we let the preceding layers predict the closest nearest neighbors. It mimics kNN because the first layer predicts the label of the first closest vector to $x$, the second layer predicts the label of the second closest vector to $x$, etc. 

This model takes a feature vector $x$, its ground truth label $Y^{GT}$, a random subset $x_1,x_2,...,x_n$ from the training set (to be stored in the external memory) and $Y_1^T,Y_2^T,...,Y_K^T$ denoting the labels of the $1^{st},2^{nd},...,K^{th}$ nearest samples to $x$ among the entire training set (calculated upfront using kNN). It predicts a label $Y^P$ and a sequence of $K$ labels of closest samples $Y_1^P,Y_2^P,...,Y_K^P$. 

After the $t^{th}$ layer, by a softmax operation with temperature after a linear mapping $(W_y,b_y)$, the model predicts the label of $t^{th}$ nearest sample by
$$Y_t^P=softmax((W_y( Hu^t+o^t)+b_y)/\tau)$$
where $t=1,2,...,K$. The role of $\tau$ is the same as in the V2LS model. Taking the average of the predicted label distributions, the final label of $x$ is calculated by
$$Y^P= \mathlarger{\sum}_{t=1}^{K}Y_t^P/K.$$

Same as V2LS, the loss function of MNkNN is defined as
$$\overline{L_1}=\displaystyle \mathop{\mathbb{E}_{}}\{\mathlarger{\sum}_{t=1}^{K}KL(Y_t^T||Y_t^P)/K+\alpha KL(Y^{GT}||Y^P)\}.$$

The first term accounts for learning neighboring information, and the second term forces the model to provide the best single candidate class.

In inference, the model takes a query $x$ and random samples $x_1,x_2,...,x_n$ from the training set, and generates the predicted label $Y^P$ as well as a sequence of nearest labels $Y_1^P,Y_2^P,...,Y_K^P$.

\subsection{Memory Network - kNN with Vector Sequence (MNkNN\_VEC) Model} 

This model is built on MNkNN, but it predicts out-of-sample feature vectors $X_t^P$ as well. MNkNN\_VEC takes a query feature vector $x$, its corresponding ground truth label $Y^{GT}$, a random subset $x_1,x_2,...,x_n$ from the training dataset (to be stored in the external memory), $Y_1^T,Y_2^T,...,Y_K^T$ and $X_1^T,X_2^T,...,X_K^T$ denoting labels and feature vectors of the $1^{st},2^{nd},...,K^{th}$ nearest samples to $x$ among the entire training set (calculated both upfront using kNN). MNkNN\_VEC predicts a label $Y^P$, a sequence of labels $Y_1^P,Y_2^P,...,Y_K^P$ and a sequence of out-of-sample feature vectors $X_1^P,X_2^P,...,X_K^P$. 

By a linear mapping $T$, a $ReLU$ operation and another linear mapping $(W_x,b_x)$, the feature vectors are then calculated by 
$$X_t^P=W_xmax\{T( Hu^t+o^t),0\}+b_x.$$

Same as the V2VSLS model, combining the $L^2$ norm and the KL divergence together, the loss function is defined as
$$\overline{L}=\displaystyle \overline{L_1}+\lambda \mathop{\mathbb{E}_{}}\{\mathlarger{\sum}_{t=1}^{K}||X_t^P-X_t^T||^2\}$$
As discussed in the V2VSLS model, having such loss function forces the model to learn both the feature vectors and the labels of nearest neighbors. 

In inference, the model takes a query $x$ and random vectors $x_1,x_2,...,x_n$ from the training dataset, and generates the final label $Y^P$, a sequence of labels $Y_1^P,Y_2^P,...,Y_K^P$ and a sequence of out-of-sample feature vectors $X_1^P,X_2^P,...,X_K^P$.

\subsection{Out-of-Core Models}

In the models exhibited so far, we assume that kNN can be run on the entire dataset exactly to compute the $K$ nearest feature vectors and corresponding labels to an input sample. However, there are two problems with this assumption. First, this can be very computationally expensive if the dataset size is large. Second, the training dataset might be too big to fit in memory. When either of these two challenges is present, an out-of-core model assuming it is infeasible to run a `full' kNN on the entire dataset has to be invoked. The out-of-core models avoid running kNN on the entire dataset, and thus save computational time and resources. 

Let $B$ be the maximum number of samples that can be stored in memory, where $B>K$. For a training sample $x$, we sample a subset $S$ from the training set (including $x$) where $|S|=B$, then we run kNN on $S$ to obtain the $K$ nearest feature vectors and corresponding labels to $x$, which are denoted as $Y^T(S)=\{Y_1^T,Y_2^T,...,Y_K^T\}$ and $X^T(S)=\{X_1^T,X_2^T,...,X_K^T\}$ for $x$ in the training process. The previously introduced loss functions $L$ and $\overline{L}$ depend on $x,Y^{GT},X^T(S),Y^T(S)$ and the model parameters $\Theta$, and thus our out-of-core models are to solve
$$
\min_{\Theta} \mathop{\mathbb{E}_{x}} \mathop{\mathbb{E}_{S}}\{\widetilde{L}(x,Y^{GT},X^T(S),Y^T(S),\Theta)\}
$$
where $\widetilde{L}$ is either $L$ or $\overline{L}$.

Sampling a set of size $B$ and then finding the nearest $K$ samples only once, however, are insufficient on imbalanced datasets, due to the low selection probability for minor classes. To resolve this, we iteratively take $R$ random batches: each time a random batch is taken, we update the closest samples $X^T(S)$ by the $K$ closest samples among the current batch and the $K$ previous closest samples. These resulting nearest feature vectors and corresponding labels are used as $X^T(S)$ and $Y^T(S)$ for $x$ in the loss function. Note that we allow the previously selected samples to be selected in later sampling iterations. The entire algorithm is exhibited in Algorithm 1. 

\begin{algorithm}[h!]
 \For{epoch = 1,...,T}{
  \For{training sample $x$}{
  Let $X^T=\varnothing$\;
   \For{r = 1 to R}{
  Randomly draw $B$ samples from training set\;
  $U=$ nearest $K$ samples to $x$ among the $B$ samples\;
  Let $X^T$ be the nearest $K$ samples to $x$ among $U \cup X^T$\;
}
 Update parameters by a gradient iteration: $\Theta^R$$=\Theta^R-\alpha\nabla \widetilde{L}(x,Y^{GT},X^T,Y^T,\Theta^R)$\;
  }
 }
 \caption{Out-of-core framework}
\end{algorithm}



\section{Computational experiments}

In this section, we evaluate our models on 9 classification datasets: Network Intrusion (NI) \cite{NI}, Forest Covertype (COV) \cite{COV}, SensIT \cite{sensit}, Credit Card Default (CCD) \cite{CCD}, MNIST\cite{mnist}, CIFAR-10\cite{cifar}, News20\cite{news20}, IMDb\cite{imdb} and Reuters \cite{reuters}. All of the models have been developed in Python 2.7 by using Tensorflow 1.4 and Pytorch 0.4.


For each dataset we experiment with 5 different seeds and all reported numbers are averages taken over 5 random seeds. We discuss the performance of the models in two aspects: classification and oversampling.

\subsection{Classification}
\textbf{Experimental Setup}
As comparisons against memory network kNN models and sequence to sequence kNN models, we use kNN with Euclidean metric and several currently best classification models random forest (RF), extreme gradient boosting (XGB), lightGBM (LGBM), a 4-layer feed-forward neural network (FFN) trained using the Adam optimization algorithm (which has been calibrated) with dropout and batch normalization and MemN2N (since MNkNN and MNkNN\_VEC are built on MemN2N) as benchmarks. Value $K=5$ is used in all models because it yields the best performance with low standard deviation among $K=1,2,...,20$. Increasing $K$ beyond $K=5$ is somewhat detrimental to the F-1 scores while significantly increasing the training time.

In the sequence to sequence kNN models, LSTM cells are used. In the memory network kNN models, the size of the external memory is 64 since we observe that models with memory vectors of size 64 generally provide the best F-1 scores with acceptable running time. Both sequence to sequence kNN models and memory network kNN models are trained using the Adam optimization algorithm with initial learning rate set to be 0.01. Dropout with probability 0.2 and batch normalization are used to avoid overfitting. Regarding the choices of other hyper-parameters, we find that $\tau=0.85$, $\lambda=0.12$ and $\alpha=9.5$ provide overall the best F-1 scores. 

We first experiment on structured datasets not requiring special embeddings, i.e. NI, COV, SensIT and CCD. Details of these datasets are in Table \ref{tab:dataset}. We only consider 3 classes in NI and COV datasets due to significant class imbalance. 

\begin{table}[h!]
\centering
\footnotesize
\caption{Datasets information.} 
  \begin{tabular}{r | r  | r | r | r }
  \hline
      & \textbf{NI} & \textbf{COV} & \textbf{SensIT}&\textbf{CCD}\\ \hline
Dataset Size  & 796,497 & 530,895 &98,528& 30,000\\\hline
Feature Size & 41 & 54 & 100&23 \\ \hline
Number of Classes & 3  &3 & 3 &2\\\hline

  \end{tabular}
  \label{tab:dataset}
\end{table}

\textbf{Overall Results of Full Model on Structured Data}
We first discuss the full models that can handle all of the training data, i.e. kNN can be run on the entire dataset. Table \ref{tab:f1_full} shows that in the full model case, V2VSLS consistently outperforms the best classification models on all four datasets. t-tests show that it significantly outperforms benchmarks at the 5\% level on all four datasets. For our kNN models, V2LS significantly outperforms V2VS, because V2VS tries to reconstruct the feature level information as explained in Section III-B, which does not utilize the label information. Moreover, it can also be seen that predicting not only labels but feature vectors as well is reasonable, since V2VSLS consistently outperforms V2LS and MNkNN\_VEC consistently outperforms MNkNN. Models predicting feature vectors outperform models not predicting feature vectors on all datasets. These memory based models exhibit subpar performance, which is expected since they only consider 64 training samples at once (despite using exact labels). 


\begin{table}[h!]
\centering
\caption {F-1 score comparison of full models.}  
\begin{tabular}{r | r  | r | r | r }\hline
       & \textbf{NI} & \textbf{COV} & \textbf{SensIT} & \textbf{CCD}\\ \hline
kNN  & 90.54 & 91.15&  82.56& 63.81\\\hline
RF& 90.44&93.76&82.70&66.94 \\\hline
XGB&87.53&91.98&82.56&66.95 \\\hline
LGBM&90.23&89.85&83.29&65.68 \\\hline
FFN  & 88.53 & 91.83&  83.67& 65.37\\\hline
MemN2N& 79.36 &77.98&75.17&61.83\\\hline
V2LS &91.28&93.94&84.93&68.38\\\hline
V2VS& 86.18 &90.39&74.84&64.23\\\hline
V2VSLS& \textbf{92.07} &\textbf{94.97}&\textbf{86.24}&\textbf{69.87}\\\hline
MNkNN& 83.83 &80.12&79.58&67.26\\\hline
MNkNN\_VEC&  84.59 & 83.94 &83.41&68.82\\\hline

  \end{tabular}
\label{tab:f1_full}
\end{table}

\textbf{Overall Results of Full Model on Unstructured Data}
To provide insights of how our model performs on unstructured data, we further evaluate our model on the following image and text datasets: MNIST, CIFAR-10, News20, IMDb and Reuters. MNIST and CIFAR-10 are two conventional datasets for image classification and News20, IMDb and Reuters are commonly used datasets for text classification.

A 5-layer convolutional neural network is trained on MNIST and the network except the top fully connected layer is used to extract features. The extracted MNIST feature size is 1,024. For CIFAR-10, we train an 18-layer ResNet \cite{resnet} and extract the CIFAR-10 features using the trained ResNet but the top fully connected layer. The extracted feature size is 1,024. To extract features of News20, we calculate the tf-idf vector for each document and apply SVD to reduce the feature dimension to 3,000. By stacking a temporal convolutional layer and a LSTM layer we obtain the 64-dimensional features for each movie review in IMDb. For Reuters we use a convolutional-LSTM network to extract 128-dimensional feature vectors.

We compare V2VSLS with some of the most popular classification models (fine-tuned) in Table \ref{table:more_experiments}.  Note that accuracy is used and FFN in Table \ref{table:more_experiments} denotes 5-layer CNN on MNIST, 18-layer ResNet on CIFAR-10, 7-layer FFN on News20 and 5-layer convolutional-LSTM network on IMDb and Reuters.

\begin{table}[h!]
\centering
\caption{Accuracy on unstructured datasets.}
  \begin{tabular}{r | r  | r | r|r |r}\hline
      & \textbf{MNIST} & \textbf{CIFAR-10} & \textbf{News20}&\textbf{IMDb}&\textbf{Reuters}\\ \hline
kNN & 98.91&93.12&62.14&86.25&72.34\\\hline
RF& 98.87&92.69&58.55&87.41&73.70 \\\hline
SVM& 98.86&92.95&71.84&87.75&73.97 \\\hline
LGBM&99.57&92.18&70.59&88.60&74.98 \\\hline
XGB& 99.12&91.55&69.81&88.51&74.24 \\\hline
FFN & 99.51&94.41&\textbf{72.93}&88.33&74.83\\\hline
V2VSLS& \textbf{99.70}&\textbf{94.86}&72.68 &\textbf{89.83}&\textbf{76.11}\\\hline
  \end{tabular}
  \label{table:more_experiments}
\end{table}

On News20, 7-layer FFN performs slightly better than V2VSLS, but V2VSLS consistently outperforms other classification models on all other datasets. The performance of V2VSLS on other datasets is close to some of the state-of-the-art deep models specifically designed for image or text data. There is a slight gap attributed to the single stage employed by pure deep learning models v.s. our experiment that has two stages (embedding construction, kNN). Nevertheless, the performance of V2VSLS on these unstructured datasets still outperforms many currently popular models.

\textbf{Overall Results of Out-of-Core Model}
Next, we validate our models in the out-of-core scenario which avoids running kNN on the entire dataset for saving computational resources. In the out-of-core versions of our models, $R$ is set to be 50, since we observe that increasing 50 to, for instance, 100, only has a slight impact on F-1 scores. However, increasing $R$ from 50 to 100 substantially increases the running time. The batch size $B$ of the out-of-core models is set to be 64 since it is found to provide overall the best F-1 scores with reasonable running time.


\begin{table}[h!]
\centering
\caption {F-1 score of out-of-core model with R=50. }  
  \begin{tabular}{r | r  | r | r | r }\hline
       & \textbf{NI} & \textbf{COV} & \textbf{SensIT} & \textbf{CCD}\\ \hline
kNN  & 73.87 & 63.87& 61.40& 59.41\\\hline
V2LS &90.63&90.29&82.47&67.51\\\hline
V2VS& 81.92 &71.29&69.12&61.36\\\hline
V2VSLS& \textbf{91.27} &\textbf{92.89}&\textbf{83.38}&\textbf{69.21}\\\hline
MNkNN& 81.89 &78.58&78.80&66.16\\\hline
MNkNN\_VEC&  83.19 & 81.72 &82.32&68.15\\\hline

  \end{tabular}
  \label{tab:ooc}

\end{table}

Table \ref{tab:ooc} shows the results of our models under the out-of-core assumption when $R=50$ and $B=64$. Both V2VLSL and MNkNN\_VEC significantly outperform the kNN benchmark based on t-tests at the 5\% significance level. The kNN benchmark provides a low score since we restrict the batch size (or memory size) to be 64, and it turns out that kNN is substantially affected by the randomness of batches. Our models (except V2VS, since it makes predictions only depending on feature vector sequences) are robust under the out-of-core setting, because the weight of the ground truth label in the loss function is relatively high so that even if the input nearest sequences are noisy, they still can focus on learning the ground truth label and making reasonable predictions. 

\begin{table}[h!]
\centering
\footnotesize
\caption{Full model and out-of-core (OOC) model comparison on SensIT.} 
    \begin{tabular}{r |r|r|r}\hline
       & \textbf{kNN} & \textbf{V2LS} & \textbf{V2VS}\\ \hline
Full F-1& 82.56&84.93&74.84\\
OOC F-1 & 61.40&82.47&69.12\\\hline
Full time (s) & 312  &  443+635 & 857+1358 \\
OOC time (s)  & 193  & 287+619  & 488+1316\\ \hline
  \end{tabular}
  \begin{tabular}{r | r  | r |r}\hline
       & \textbf{V2VSLS}&\textbf{MNkNN}&\textbf{MNkNN\_VEC}\\ \hline
Full F-1&86.24&79.58&83.41\\
OOC F-1 & 83.38&78.80&82.32\\\hline
Full time (s) &1391+1802 & 443+692& 1391+1081 \\
OOC time (s)  & 741+1846& 287+703& 741+1055\\ \hline
  \end{tabular}
\label{tab:full_ooc_comp}
\end{table}

\textbf{Full Model and Out-of-Core Model Comparison}
Table \ref{tab:full_ooc_comp} shows a comparison between the full and out-of-core models with $R=50, B=64$ on the SensIT dataset. The running time of our models are broken down to two parts: the first part is the time to obtain sequences of $K$ nearest feature vectors and labels and the second part is the model training time. Under the out-of-core setting, overall the kNN sequence preprocessing time is saved by approximately 40\% while the models perform only slightly worse.

In the next paragraphs, we focus the discussion on full models without the out-of-core assumption, in order to achieve the best classification performance.

\textbf{Comparison with a Set-Based Model and Swapped Order V2VSLS} We evaluate the necessity of modeling nearest neighbors as a sequence, instead of as a set. First, we compare the set-based model with the V2VSLS model. Note that compared to V2VSLS, the set-based neural network model also predicts $K$ labels, $K$ nearest neighbors and a final label for classification, but it does not model the $K$ labels and nearest neighbors as a sequence. The only difference between the set-based model and V2VSLS is that the set-based model's outputs are orderless. As shown in Table \ref{tab:f1_skipgram}, the set-based model which removes the order of nearest neighbors suffers from a consistent performance drop across datasets. The set-based model still outperforms most of the existing popular classification methods, however, which again validates that predicting nearest neighbors is beneficial for classification.

\begin{table}[h!]
\centering
\footnotesize
\caption{F-1 score of the set-based model, V2VSLS with arbitrarily swapped nearest neighbors and original V2VSLS.}
  \begin{tabular}{r | r  | r | r | r }\hline
      & \textbf{NI} & \textbf{COV} & \textbf{SensIT}&\textbf{CCD}\\ \hline
Set-based model & 91.25&94.10&85.51&68.77 \\\hline
Swapped order V2VSLS& 91.79&94.56&85.99&69.42 \\\hline
V2VSLS & \textbf{92.07}&\textbf{94.97}&\textbf{86.24}&\textbf{69.87}\\\hline

  \end{tabular}
  \label{tab:f1_skipgram}
\end{table}

Following the orderless nearest neighbors experiment, to get a better understanding on the order of nearest neighbors, we arbitrarily swap the first and the third nearest neighbor of the order in the training data. Intuitively, if the performance drops after swapping the nearest neighbors in the training data, utilizing the order information of nearest neighbors is crucial. The results are shown in Table \ref{tab:f1_skipgram}. V2VSLS with swapped order performs worse than V2VSLS with original order, but it still outperforms the set-based model consistently, which validates that keeping the order of the nearest neighbors is necessary. 

\textbf{Comparison with other Related Benchmark Models}
We first compare V2VSLS with the popular ranking-based model LambdaRank \cite{lambdarank} with lightGBM as the classifier. LambdaRank ranks the feature vectors in the training set for a given query by similarity. In inference, for a provided feature vector query, LambdaRank ranks the feature vectors in the training set and obtains the nearest $K=5$ samples. Finally, the label of the query is obtained by majority voting among the corresponding labels of those nearest samples in the training set. The results are provided in Table \ref{tab:f1_lmdrk}. The performance improvement of approximately 40\% is observed in the experiment, which shows the significant superiority of V2VSLS over the conventional ranking-based method.

\begin{table}[h!]
\centering
\footnotesize
\caption{F-1 score comparison with LambdaRank \cite{lambdarank}.}
  \begin{tabular}{r | r  | r | r | r }\hline
      & \textbf{NI} & \textbf{COV} & \textbf{SensIT}&\textbf{CCD}\\ \hline
LambdaRank & 45.58&59.81&41.03&38.91 \\\hline
kNN-Augmented Networks& 64.18&69.64&54.29&52.18 \\\hline
MemN2N& 79.36&77.98&75.17&61.83 \\\hline
MNkNN\_VEC& 84.59&83.94&83.41&68.82 \\\hline
V2VSLS  & \textbf{92.07}&\textbf{94.97}&\textbf{86.24}&\textbf{69.87}\\\hline
  \end{tabular}
  \label{tab:f1_lmdrk}
\end{table}

To compare our work with another similar work which utilizes the nearest neighbors to make predictions, we implemented the kNN-Augmented Networks from \cite{knntext}. The comparison is shown in Table \ref{tab:f1_lmdrk}. In the implementation of kNN-Augmented Networks, we have fine-tuned the hyper-parameters: $K=5$, $I=8$, learning rate=$0.001$ and the Adam optimizer. There is a substantial gap between our models and the kNN-Augmented Networks, that we were unable to close despite a significant effort to fine tune the hyper-parameters.

\subsection{Oversampling}

\textbf{Experimental Setup}
When class distributions are highly imbalanced, many classification models have low accuracy or F-1 score on the minority class. A simple but effective way to handle this problem is to oversample the minority class. Since V2VSLS and MNkNN\_VEC are able to predict out-of-sample feature vectors, we also regard our models as oversamplers and we compare them with two widely used oversampling techniques: SMOTE and ADASYN. We only test V2VSLS since it is the best model that can handle all of the data. In our experiments, we evaluate our model on four imbalanced datasets. We first fully train the model, and then for each sample from the training set, V2VSLS predicts $K=5$ out-of-sample feature vectors which are regarded as synthetic samples. We add them to the training set if they are in a minority class until the classes are balanced or there are no minority training data left for creating synthetic samples. In our oversampling experiments, we use $\lambda=1.3$ and $\alpha=3$.

\begin{table}[h!]
\centering
\caption {Oversampling: F-1 score comparison.  }    \begin{tabular}{r | r  | r | r | r }\hline
       & \textbf{NI} & \textbf{COV} & \textbf{SensIT} & \textbf{CCD}\\ \hline
FFN-original  & 89.64 & 91.83& 83.67&65.37\\
FFN-SMOTE &89.99&91.18&83.43&66.32\\
FFN-ADASYN& 90.38 &90.67&83.72&66.51\\
FFN-V2VSLS& \textbf{90.89} &\textbf{92.05}&\textbf{83.94}&\textbf{66.82}\\\hline
XGB-original  & 87.53 & 91.98& 82.56& 66.95\\
XGB-SMOTE &87.79&91.86&82.87&66.56\\
XGB-ADASYN& \textbf{88.39} &\textbf{92.56}&\textbf{83.42}&66.20\\
XGB-V2VSLS& 87.62 &92.43&82.46&\textbf{66.96}\\\hline
RF-original  & 90.44 & 93.76& 82.70& 66.94\\
RF-SMOTE &89.97&93.88&83.01&66.13\\
RF-ADASYN& 89.39 &93.83&\textbf{83.34}&67.14\\
RF-V2VSLS& \textbf{90.79} &\textbf{94.36}&82.75&\textbf{68.08}\\\hline
  \end{tabular}
  \label{tab:oversample}
\end{table}


\textbf{Overall Results on Oversampling}
Table \ref{tab:oversample} shows the F-1 scores of FFN, extreme gradient boosting and random forest classification models, with different oversampling techniques, namely, original training set without oversampling, SMOTE, ADASYN and V2VSLS. V2VSLS performs the best among all combinations of classification models and oversampling techniques, as shown in Table \ref{tab:oversample_comp}. Although most of the time models on datasets with three oversampling techniques outperform models on datasets without oversampling, the classification performance still largely depends on the classification model used and which dataset is considered.

\begin{table}[h!]
\centering
\caption{Oversampling techniques comparison.}
  \begin{tabular}{r | r  | r | r |r}\hline
       & \textbf{NI} & \textbf{COV} & \textbf{SensIT}&\textbf{CCD}\\ \hline
       Best model& FFN+V2VSLS &RF+V2VSLS & FFN+V2VSLS&RF+V2VSLS\\\hline
Best F-1 & 90.89 & 94.36 &83.92& 68.08\\\hline
Better than best SMOTE & 0.51\% &0.61\%& 2.28\%& 1\%\\\hline
Better than best ADASYN& 0.6\% & 0.56\% &0.26\%& 1.4\%\\\hline
  \end{tabular}
  \label{tab:oversample_comp}
\end{table}

\begin{figure}[h!]
    \includegraphics[width=0.48\textwidth]{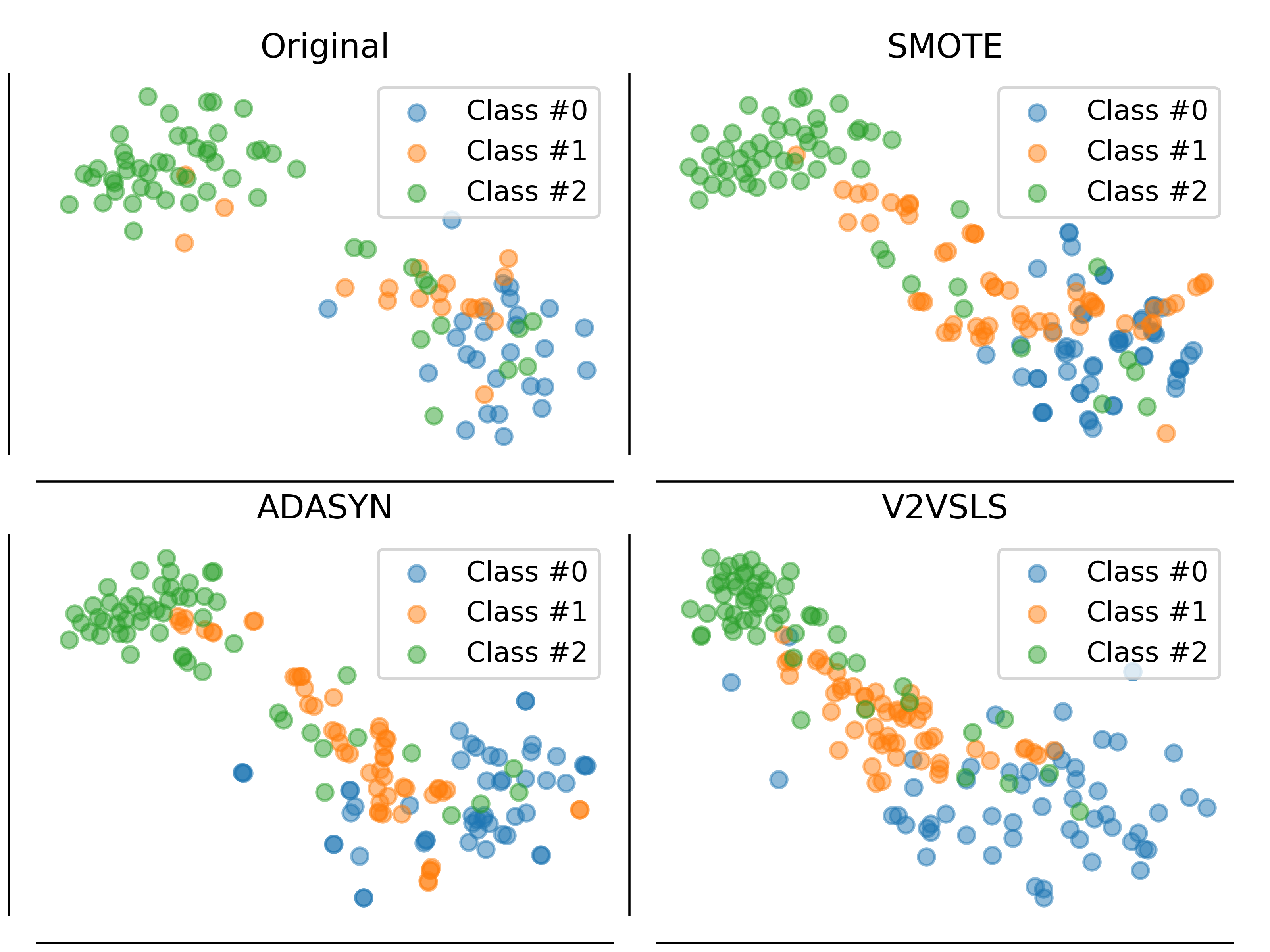}
    \centering
  \caption{t-SNE visualization of different oversampling methods}
  \centering
  \label{fig:tsne}
  \end{figure}

\textbf{Synthetic Samples Visualization}
Figure \ref{fig:tsne} shows a t-SNE \cite{tsne} visualization of the original set and the oversampled set, using SensIT dataset, projected onto 2-D space. Although SMOTE and ADASYN overall perform well, their class boundaries are not as clean as those obtained by V2VSLS, which validates the effectiveness of V2VSLS oversampler. \\

\section{Conclusion}

In summary, we find that it is beneficial to have neural network models learn not only labels but feature vectors as well. In the full models, V2VSLS outperforms all other models consistently and performs well on unstructured data; in the out-of-core models, both V2VSLS and MNkNN\_VEC significantly outperform the kNN benchmark. As an oversampler, the average F-1 score based on the training set augmented by V2VSLS outperforms that of SMOTE and ADASYN. 

We recommend to run V2VSLS with large $\alpha$ and small $\lambda$ for classification. In the oversampling scenario, however, we suggest to use small $\alpha$ and large $\lambda$ so that the model focuses more on the feature vectors, i.e. synthetic samples. 

\bibliography{IEEEexample}

\begin{thebibliography}{10}
\providecommand{\url}[1]{#1}
\csname url@samestyle\endcsname
\providecommand{\newblock}{\relax}
\providecommand{\bibinfo}[2]{#2}
\providecommand{\BIBentrySTDinterwordspacing}{\spaceskip=0pt\relax}
\providecommand{\BIBentryALTinterwordstretchfactor}{4}
\providecommand{\BIBentryALTinterwordspacing}{\spaceskip=\fontdimen2\font plus
\BIBentryALTinterwordstretchfactor\fontdimen3\font minus
  \fontdimen4\font\relax}
\providecommand{\BIBforeignlanguage}[2]{{%
\expandafter\ifx\csname l@#1\endcsname\relax
\typeout{** WARNING: IEEEtranS.bst: No hyphenation pattern has been}%
\typeout{** loaded for the language `#1'. Using the pattern for}%
\typeout{** the default language instead.}%
\else
\language=\csname l@#1\endcsname
\fi
#2}}
\providecommand{\BIBdecl}{\relax}
\BIBdecl

\bibitem{att1}
D.~Bahdanau, K.~Cho, and Y.~Bengio, ``Neural machine translation by jointly
  learning to align and translate,'' \emph{International Conference on Learning
  Representations}, 2015.

\bibitem{COV}
\BIBentryALTinterwordspacing
J.~A. Blackard and D.~J. Dean, ``Forest covertype,'' \emph{{UCI} Machine
  Learning Repository}, 1998. [Online]. Available:
  \url{https://archive.ics.uci.edu/ml/datasets/covertype}
\BIBentrySTDinterwordspacing

\bibitem{RF}
L.~Breiman, ``Random forests,'' \emph{Machine Learning}, 2001.

\bibitem{lambdarank}
C.~J.~C. Burges, R.~Ragno, and Q.~V. Le, ``Learning to rank with nonsmooth cost
  functions,'' in \emph{Conference on Neural Information Processing Systems},
  2006.

\bibitem{ir5}
G.~Carlsson, T.~Ishkhanov, V.~de~Silva, and A.~Zomorodian, ``On the local
  behavior of spaces of natural images,'' \emph{International Journal of
  Computer Vision}, 2008.

\bibitem{smote}
N.~V. Chawla, K.~W. Bowyer, L.~O. Hall, and W.~P. Kegelmeyer, ``{SMOTE}:
  Synthetic minority over-sampling technique,'' \emph{Journal of Artificial
  Intelligence Research}, 2002.

\bibitem{XGB}
T.~Chen and C.~Guestrin, ``Xgboost: A scalable tree boosting system,''
  \emph{ACM SIGKDD International Conference on Knowledge Discovery and Data
  Mining}, 2016.

\bibitem{gru}
K.~Cho, B.~van Merrienboer, {\c{C}}.~G{\"{u}}l{\c{c}}ehre, F.~Bougares,
  H.~Schwenk, and Y.~Bengio, ``Learning phrase representations using {RNN}
  encoder-decoder for statistical machine translation,'' \emph{Conference on
  Empirical Methods in Natural Language Processing}, 2014.

\bibitem{reuters}
\BIBentryALTinterwordspacing
F.~Chollet, ``Reuters newswire topics classification,'' 2015. [Online].
  Available: \url{https://keras.io/datasets}
\BIBentrySTDinterwordspacing

\bibitem{knn}
T.~M. Cover and P.~E. Hart, ``Nearest neighbor pattern classification,''
  \emph{Institute of Electrical and Electronics Engineers Transactions on
  Information Theory}, 1967.

\bibitem{ir2}
J.~Deng, W.~Dong, R.~Socher, L.-J. Li, K.~Li, and L.~Fei-Fei, ``Imagenet: A
  large-scale hierarchical image database,'' \emph{Conference on Computer
  Vision and Pattern Recognition}, 2009.

\bibitem{sensit}
M.~F. Duarte and Y.~H. Hu, ``Vehicle classification in distributed sensor
  networks,'' \emph{Journal of Parallel and Distributed Computing}, 2004.

\bibitem{ae1}
D.~Erhan, Y.~Bengio, A.~C. Courville, P.-A. Manzagol, P.~Vincent, and
  S.~Bengio, ``Why does unsupervised pre-training help deep learning?''
  \emph{Journal of Machine Learning Research}, 2010.

\bibitem{adasyn}
H.~He, Y.~Bai, E.~A. Garcia, and S.~Li, ``Adasyn: Adaptive synthetic sampling
  approach for imbalanced learning,'' \emph{International Joint Conference on
  Neural Networks}, 2008.

\bibitem{ae2}
K.~He, X.~Zhang, S.~Ren, and J.~Sun, ``Deep residual learning for image
  recognition,'' \emph{Conference on Computer Vision and Pattern Recognition},
  2016.

\bibitem{resnet}
------, ``Deep residual learning for image recognition,'' \emph{2016 IEEE
  Conference on Computer Vision and Pattern Recognition}, pp. 770--778, 2016.

\bibitem{NI}
\BIBentryALTinterwordspacing
S.~Hettich and S.~D. Bay, ``Network intrusion,'' \emph{The {UCI} {KDD}
  Archive}, 1999. [Online]. Available: \url{http://kdd.ics.uci.edu}
\BIBentrySTDinterwordspacing

\bibitem{ae4}
G.~Hinton and R.~Salakhutdinov, ``Reducing the dimensionality of data with
  neural networks,'' \emph{Science}, 2006.

\bibitem{temperature}
G.~Hinton, O.~Vinyals, and J.~Dean, ``Distilling the knowledge in a neural
  network,'' \emph{Annual Conference on Neural Information Processing Systems},
  2015.

\bibitem{lstm2}
S.~Hochreiter and J.~Schmidhuber, ``Long short-term memory,'' \emph{Neural
  Computation}, 1997.

\bibitem{karp}
\BIBentryALTinterwordspacing
A.~Karpathy, ``The unreasonable effectiveness of recurrent neural networks,''
  2015. [Online]. Available:
  \url{http://karpathy.github.io/2015/05/21/rnn-effectiveness/}
\BIBentrySTDinterwordspacing

\bibitem{lgbm}
G.~Ke, Q.~Meng, T.~Finley, T.~Wang, W.~Chen, W.~Ma, Q.~Ye, and T.~M. Liu,
  ``Lightgbm: A highly efficient gradient boosting decision tree,'' 2017.

\bibitem{cifar}
\BIBentryALTinterwordspacing
A.~Krizhevsky, V.~Nair, and G.~Hinton, ``Cifar-10 (canadian institute for
  advanced research).'' [Online]. Available:
  \url{http://www.cs.toronto.edu/~kriz/cifar.html}
\BIBentrySTDinterwordspacing

\bibitem{ir1}
A.~Krizhevsky, I.~Sutskever, and G.~E. Hinton, ``Imagenet classification with
  deep convolutional neural networks,'' \emph{Annual Conference on Neural
  Information Processing Systems}, 2012.

\bibitem{news20}
K.~Lang, ``Newsweeder: Learning to filter netnews,'' \emph{Proceedings of the
  Twelfth International Conference on Machine Learning}, pp. 331--339, 1995.

\bibitem{mnist}
\BIBentryALTinterwordspacing
Y.~LeCun and C.~Cortes, ``{MNIST} handwritten digit database,'' 1999. [Online].
  Available: \url{http://yann.lecun.com/exdb/mnist/}
\BIBentrySTDinterwordspacing

\bibitem{imdb}
A.~L. Maas, R.~E. Daly, P.~T. Pham, D.~Huang, A.~Y. Ng, and C.~Potts,
  ``Learning word vectors for sentiment analysis,'' pp. 142--150, 2011.

\bibitem{bf}
C.~Mathy, N.~Derbinsky, J.~Bento, J.~Rosenthal, and J.~S. Yedidia, ``The
  boundary forest algorithm for online supervised and unsupervised learning,''
  \emph{Association for the Advancement of Artificial Intelligence Conference},
  2015.

\bibitem{nlp2}
T.~Mikolov, W.~tau Yih, and G.~Zweig, ``Linguistic regularities in continuous
  space word representations,'' \emph{Annual Conference of the North American
  Chapter of the Association for Computational Linguistics: Human Language
  Technologies}, 2013.

\bibitem{colah}
\BIBentryALTinterwordspacing
C.~Olah, ``Neural networks, manifolds, and topology,'' 2014. [Online].
  Available: \url{http://colah.github.io/posts/2014-03-NN-Manifolds-Topology/}
\BIBentrySTDinterwordspacing

\bibitem{att2}
T.~Rockt{\"{a}}schel, E.~Grefenstette, K.~M. Hermann, T.~Kocisk{\'{y}}, and
  P.~Blunsom, ``Reasoning about entailment with neural attention,''
  \emph{International Conference on Learning Representations}, 2016.

\bibitem{MN15}
S.~Sukhbaatar, A.~Szlam, J.~Weston, and R.~Fergus, ``Weakly supervised memory
  networks,'' \emph{Annual Conference on Neural Information Processing
  Systems}, 2015.

\bibitem{s2s1}
I.~Sutskever, O.~Vinyals, and Q.~V. Le, ``Sequence to sequence learning with
  neural networks,'' \emph{Annual Conference on Neural Information Processing
  Systems}, 2014.

\bibitem{nlp1}
J.~Turian, L.~Ratinov, and Y.~Bengio, ``Word representations: A simple and
  general method for semi-supervised learning,'' \emph{Annual Meeting of the
  Association for Computational Linguistics}, 2010.

\bibitem{tsne}
L.~van~der Maaten and G.~Hinton, ``Visualizing data using {t-SNE},''
  \emph{Journal of Machine Learning Research}, 2008.

\bibitem{ae3}
P.~Vincent, H.~Larochelle, I.~Lajoie, Y.~Bengio, and P.~antoine Manzagol,
  ``Stacked denoising autoencoders: learning useful representations in a deep
  network with a local denoising criterion,'' \emph{Journal of Machine Learning
  Research}, 2010.

\bibitem{knntext}
\BIBentryALTinterwordspacing
Z.~Wang, W.~Hamza, and L.~Song, ``k-nearest neighbor augmented neural networks
  for text classification,'' \emph{arXiv Repository}, 2017. [Online].
  Available: \url{http://arxiv.org/abs/1708.07863}
\BIBentrySTDinterwordspacing

\bibitem{MN14}
J.~Weston, S.~Chopra, and A.~Bordes, ``Memory networks,'' \emph{International
  Conference on Learning Representations}, 2015.

\bibitem{CCD}
I.-C. Yeh and C.~hui Lien, ``The comparisons of data mining techniques for the
  predictive accuracy of probability of default of credit card clients.''
  \emph{Expert Systems with Applications}, 2009.

\bibitem{dbt}
\BIBentryALTinterwordspacing
D.~Zoran, B.~Lakshminarayanan, and C.~Blundell, ``Learning deep nearest
  neighbor representations using differentiable boundary trees,'' \emph{arXiv
  Repository}, 2017. [Online]. Available: \url{http://arxiv.org/abs/1702.08833}
\BIBentrySTDinterwordspacing

\end{thebibliography}
\bibliographystyle{IEEEtranS} 
\end{document}